\documentclass{article}
\usepackage{placeins}

\PassOptionsToPackage{numbers, compress}{natbib}
  \usepackage[preprint]{neurips_2026}

\usepackage{graphicx}
\usepackage{wrapfig}

\usepackage[utf8]{inputenc} 
\usepackage[T1]{fontenc}    
\usepackage{hyperref}       
\usepackage{url}            
\usepackage{booktabs}       
\usepackage{amsfonts}       
\usepackage{nicefrac}       
\usepackage{microtype}      
\usepackage{xcolor}         

\usepackage{amsmath, amssymb, amsthm, amsfonts}
\usepackage{bm}
\usepackage{mathtools}
\usepackage{booktabs}
\usepackage{multirow}
\usepackage{array}
\usepackage{graphicx}
\usepackage{caption}
\usepackage{subcaption}
\usepackage{xcolor}
\usepackage{tabularx}
\usepackage{makecell}
\usepackage{graphicx}
\usepackage{hyperref}
\usepackage{url}
\usepackage{xcolor}
\usepackage{verbatim}
\usepackage{algorithm}
\usepackage{algorithmic}

\newtheorem{proposition}{Proposition}
\newcommand{\Dalign}{D_{\text{align70}}}
\newcommand{\Dours}{D_{\text{ours}}}

\newcommand{\dmean}{\mu}

\definecolor{daeuncolor}{RGB}{0,102,204}

\title{What Does the Caption \emph{Really} Say? \\
Counterfactual Phrase Intervention for Compositional \\
Data Selection in Vision-Language Pretraining}

\author{%
  Hyejin Go \quad
  Semi Lee \quad
  Hyesong Choi\thanks{Corresponding author.} \\
  Soongsil University \\
  Seoul, Republic of Korea \\
  \texttt{hyejin1612@soongsil.ac.kr} \quad
  \texttt{seemi.tech@gmail.com} \quad
  \texttt{hyesong@ssu.ac.kr}
}

\begin{document}

\maketitle

\begin{abstract} CLIP-style contrastive pretraining typically curates web-scale image-text pairs
using sample-level filtering signals, often based on pair-level alignment. We show that this signal saturates: once coarse mismatches are removed, stricter global filtering no longer tracks the compositional supervision provided by the retained captions. The reason is structural---a global score conflates whether a pair is broadly plausible with whether the individual object, attribute, and relation phrases inside the caption materially support the image-text match. The latter is what compositional generalization demands, yet pair-level filters are blind to it. We address this with \textbf{Counterfactual Phrase Intervention (CPI)}, a phrase-level curation framework that converts controlled nonce-token substitutions into image-conditioned phrase-sensitivity scores. CPI uses global alignment only for coarse mismatch removal, then ranks the surviving pool by whether caption phrases measurably affect the image-text score under controlled substitution. We frame CPI as a first-order phrase-sensitivity signal rather than a grounding or identification result, and evaluate it at CC3M scale. Ranking by this signal yields a 50\%-data subset that improves VL-CheckList-VG Relation by \textbf{+1.91} over the full-data baseline and \textbf{+1.00} over alignment-only filtering at matched budget, while improving SugarCrepe overall and preserving general transfer. CPI is \emph{loss-orthogonal}: applied unchanged to NegCLIP, it further improves VL-CheckList-VG Relation by \textbf{+3.84}, with additional CE-CLIP gains in the main text. \end{abstract}
\section{Introduction}
\label{sec:intro}

CLIP-style pretraining assumes not only that a caption is globally compatible 
with an image, but that its salient objects, attributes, and relations are 
visually supported and therefore provide useful supervision. When this 
assumption fails---when a high-similarity pair is sustained by a single salient 
object, an OCR fragment, or an inadvertent text prior---the model receives a 
globally consistent yet compositionally hollow training example. Such examples 
are precisely the ones that compositionality benchmarks expose
\citep{hsieh2023sugarcrepe, dumpala2024sugarcrepe}.

The prevailing data-curation paradigm often reduces caption fidelity to a
sample-level retention score. CLIPScore-style filters
\citep{hessel2021clipscore} and related data-filtering systems---including
DataComp \citep{gadre2023datacomp}, T-MARS \citep{maini2023tmars}, DFN
\citep{fang2023data}, and Sieve \citep{mahmoud2024sieve}---refine how
image-text pairs are scored or selected. Such signals can determine whether an
image-caption pair is broadly plausible or useful under a particular filtering
pipeline, but they do not directly ask whether the match is distributed across
objects, attributes, and relations, or whether many phrases are redundant given
a dominant cue. After such coarse filtering saturates, what remains is a
\emph{support-sensitivity} problem: which phrases inside a globally matched
caption measurably affect the image-text similarity under controlled
substitution?

We address this with \textbf{Counterfactual Phrase Intervention (CPI)}, a 
phrase-level curation framework that converts controlled nonce-token 
substitutions into image-conditioned phrase-sensitivity scores. For each 
candidate phrase $p_j$ in caption $T$, CPI replaces the nominal head with a 
deterministic nonce token to obtain $\tilde{T}_j$, and measures the similarity 
drop $\Delta_j=s(I,T)-s(I,\tilde{T}_j)$. A large $\Delta_j$ indicates that the 
fixed image-text scorer is sensitive to the phrase under this controlled 
substitution; a vanishing $\Delta_j$ indicates that the phrase is comparatively 
dispensable for the measured score. Aggregating these \emph{phrase-attribution 
scores} (PAS) yields a sample-level signal that pair-level alignment cannot 
observe.

To make $\Delta_j$ meaningful, CPI uses a \emph{Three-Invariance Replacement 
Protocol}: replacements preserve exact CLIP-BPE subtoken count and surface 
syntactic form while removing lexical semantic content. These constraints 
reduce tokenization, semantic-substitution, and surface-form confounds, 
allowing $\Delta_j$ to estimate first-order sensitivity of the image-text 
similarity to the controlled replacement, up to residual text-encoder artifacts 
and higher-order phrase-interaction terms 
(Proposition~\ref{prop:controlled_sensitivity}). CPI should therefore be read 
as a controlled-substitution sensitivity signal, not as a grounding or causal 
identification method.

Operationally, CPI is deployed in a two-stage coarse-to-fine pipeline: Stage~1 
uses global alignment only to remove coarse mismatches, and Stage~2 applies 
CPI inside the surviving high-alignment pool to form a 50\% subset. We 
evaluate this subset on CC3M-scale CLIP ViT-B/32 pretraining as a controlled 
testbed, apply it unchanged to NegCLIP and CE-CLIP, and test on zero-shot, 
retrieval, linear-probe, SugarCrepe, SugarCrepe++, and VL-CheckList-VG 
benchmarks. Scaling to CC12M, DataComp, or LAION-scale pretraining is left to 
future work.

This framing distinguishes CPI from objective-side compositional VLM methods, 
which construct harder negatives or modify losses. CPI instead intervenes on 
the positive data distribution: before asking a loss to enforce finer 
compositional distinctions, it asks whether the positive pairs contain 
phrase-level supervision worth learning from. By \emph{loss-orthogonal}, we 
mean the curated subset is applied without any modification to the training 
objective, optimizer, or architecture; only the training data identities 
differ from the full-data baseline.

\paragraph{Contributions.}
\begin{enumerate}
    \item \textbf{Conceptual.} We expose a residual failure mode of pair-level
    curation: globally aligned captions can remain compositionally hollow 
    because their constituent phrases need not measurably support the 
    image-text match.

    \item \textbf{Methodological.} We introduce the
    \emph{Three-Invariance Replacement Protocol}, a deterministic nonce-based
    controlled-substitution procedure that preserves subtoken count and 
    surface syntactic form while removing lexical semantic content. Unlike prior
    controlled-substitution probes used primarily for diagnosis, CPI
    operationalizes nonce interventions as an offline data-curation signal,
    attaining 99.79\% valid-perturbation coverage.

    \item \textbf{Empirical.} On CC3M-derived CLIP pretraining, CPI improves
    phrase-sensitive metrics at a matched 50\% data budget, with the clearest
    gains on VL-CheckList-VG Relation and SugarCrepe overall, while preserving
    general transfer.

    \item \textbf{Framework.} We position phrase-level data selection as a
    loss-orthogonal data-side intervention. The same CPI-curated subset plugs 
    into NegCLIP and CE-CLIP without objective modification, providing evidence 
    that data-side phrase sensitivity can complement objective-side 
    compositional training.
\end{enumerate}
\section{Related Work}
\label{sec:related}

\paragraph{Data filtering for vision-language pretraining.}
Data curation has become a central lever for improving CLIP-style pretraining.
CLIPScore~\citep{hessel2021clipscore} and related data-filtering systems---including
DataComp~\citep{gadre2023datacomp}, T-MARS~\citep{maini2023tmars},
DFN~\citep{fang2023data}, MetaCLIP~\citep{xu2024demystifying},
Sieve~\citep{mahmoud2024sieve}, and recent active curation approaches such as
ACID/ACED~\citep{udandarao2025active}---score or select image-text pairs using
signals such as pretrained similarity, learned quality scores, OCR-aware
adjustments, text-quality estimates, or online batch utility. Other
subset-selection methods exploit dataset-level structure, including redundancy
reduction, coverage preservation, and loss- or utility-based sampling
\citep{tirumala2023d4, abbas2023semdedup, mindermann2022prioritized,
evans2024bad}. Despite these differences, many filtering methods ultimately
assign each candidate pair a sample-level retention score. They can determine
whether an image-caption pair is broadly plausible or useful under a particular
filtering pipeline, but they do not directly ask how that plausibility is
distributed across the caption's constituent phrases. CPI instead estimates
phrase-level sensitivity under controlled substitution within each caption and
uses the aggregated sensitivity scores as a curation signal.

\paragraph{Compositional vision-language learning.}
Compositional VLM methods typically intervene on the model or training objective.
NegCLIP \citep{yuksekgonul2023when}, CE-CLIP \citep{zhang2024ce},
FSC-CLIP \citep{oh2024fsc}, TripletCLIP \citep{patel2024triplet},
CLIC \citep{doveh2024dense}, and READ \citep{sahin2024read} improve
sensitivity to objects, attributes, relations, or word order through hard
negatives, reconstruction losses, sentence-level alignment, or
parameter-efficient adaptation. These approaches treat compositionality
primarily as an objective- or model-side problem. Our work is complementary:
rather than constructing harder negatives or changing the loss, CPI asks whether
the positive image-caption pairs themselves contain phrase-level supervision
worth learning from. This positions phrase-level data selection as a
loss-orthogonal data-side axis that can be combined with objective-side
compositional training.

\paragraph{Counterfactual probing.}
Counterfactual and attribution methods have long been used to analyze model
predictions, including concept-, gradient-, vocabulary-, and mediation-style
interventions
\citep{kim2018interpretability, sundararajan2017axiomatic,
oikarinen2023clip, meng2022locating}. Controlled lexical substitutions have
also been used to probe linguistic compositionality \citep{ettinger2020what}.
However, such substitutions are typically used for post hoc interpretation
rather than for constructing pretraining subsets. CPI operationalizes controlled
phrase substitution as a data-curation signal: the substitution is not merely
diagnostic, but determines which positive pairs are retained for
vision-language pretraining. 

\paragraph{Gap.}
Prior work separately studies pair-level data filtering, objective-side
compositional training, and counterfactual probing. CPI occupies their
intersection, but its role is specifically data-side: it uses phrase-level
controlled-substitution sensitivity to select positive pairs with stronger
phrase-level support, improving compositional supervision without changing the
architecture or loss. This makes CPI a complement to hard-negative and
objective-side methods rather than a replacement for them.

\paragraph{Baseline scope.}
We use alignment-only filtering as the primary matched-budget baseline to
isolate the contribution of Stage-2 phrase-level sensitivity: the competing
selection rule differs only in whether the globally aligned pool is further
ranked by CPI. This controlled comparison keeps the corpus, objective,
architecture, and data budget fixed, making it the most direct test of the
proposed curation signal. Broader comparisons with filtering systems such as
T-MARS~\citep{maini2023tmars}, DFN~\citep{fang2023data}, and
Sieve~\citep{mahmoud2024sieve} involve additional design choices, including
caption rewriting, OCR-aware suppression, learned filter networks, compute
budgets, corpora, and operating points; we therefore treat them as
complementary large-scale filtering comparisons.
\section{Motivation: The Pair-Level Saturation Phenomenon}
\label{sec:motivation}

Before introducing our framework, we document the empirical phenomenon that
motivates it. Pair-level alignment filtering is often treated as a monotonic
curation signal: stricter thresholds are expected to retain higher-quality pairs
and yield better downstream models. In contrast to the coarse filtering successes
of DataComp-style pipelines, our focus is the residual regime after global
mismatches have been removed: high pair-level alignment need not imply
phrase-level compositional support. On CC3M-scale pretraining, however, we
observe a different pattern.
 
\subsection{Alignment pruning sweep}
\label{sec:motivation:sweep}
 
We sweep the global alignment retention ratio
$\rho \in \{1.0, 0.9, 0.8, 0.7, 0.6, 0.5\}$ and pretrain CLIP ViT-B/32 on each
retained subset under identical optimization. Table~\ref{tab:align_sweep}
reports a single-seed exploratory sweep used to identify the pair-level
saturation regime.
 
\begin{table}[t]
\centering
\caption{
Alignment-only filtering sweep on CC3M used to select the Stage-1 operating
point. All models share architecture, initialization, optimizer, and step
budget. Bold marks the best per-column. This diagnostic sweep is reported
separately from the repeated-seed main comparisons in
Tables~\ref{tab:main_general} and~\ref{tab:main_compositional}.
}
\label{tab:align_sweep}
\footnotesize
\setlength{\tabcolsep}{4pt}
\begin{tabular}{lcccccccccc}
\toprule
Method & Keep & IN-1K & F-T2I R@1 & F-I2T R@1 & C10 ZS & C100 ZS & SC R/A & SC Ovr & SC++ Rep & SC++ Ovr \\
\midrule
Full   & 100\% & 7.413 & 10.260 & 14.05 & 45.685 & 14.100 & 58.63 & 57.69 & 46.81 & 45.32 \\
top90  & 90\%  & 7.492 & 10.700 & 13.500 & 44.880 & \textbf{16.120} & 58.81 & 57.86 & 47.46 & 45.64 \\
top80  & 80\%  & 7.458 & 10.700 & \textbf{15.300} & \textbf{46.850} & 14.950 & 58.99 & 57.82 & 47.06 & 45.36 \\
top70  & 70\%  & 7.558 & \textbf{11.020} & 14.200 & 45.730 & 16.070 & \textbf{59.20} & \textbf{58.34} & \textbf{47.82} & \textbf{45.95} \\
top60  & 60\%  & \textbf{7.592} & 10.560 & 13.500 & 43.740 & 14.760 & 59.43 & 58.10 & 46.92 & 45.72 \\
top50  & 50\%  & 7.477 & 10.225 & 13.100 & 42.490 & 14.100 & 58.70 & 57.92 & 46.65 & 45.69 \\
\bottomrule
\end{tabular}
\end{table}
 
Two observations are salient. First, alignment filtering is a useful coarse
signal: every retained subset, including the most aggressive top50, exceeds the
full-data baseline on SugarCrepe overall. Second, the gains do \emph{not}
increase monotonically with stricter thresholds. SugarCrepe overall peaks at
top70 (58.34) and declines thereafter; SugarCrepe++ Replace peaks at top70
(47.82) and falls to 46.65 at top50; CIFAR10 zero-shot drops from 46.85 at top80
to 42.49 at top50. This non-monotonicity appears across compositional,
retrieval, and zero-shot axes, suggesting that global alignment becomes an
incomplete ranking signal after coarse mismatches have been removed.
This trend motivates using top70 as the Stage-1 pool: it preserves the benefits
of coarse alignment filtering while leaving room for phrase-level refinement.
 
\subsection{Interpretation}
\label{sec:motivation:interpretation}

The non-monotonicity suggests that alignment filtering is most useful as a
coarse mismatch-removal signal. At moderate retention ratios, it performs its
intended function by removing globally inconsistent pairs. Beyond a saturation
point, however, the residual pool consists largely of already plausible pairs,
so further pruning by the same scalar increasingly discriminates among samples
that global alignment may not distinguish meaningfully. This makes the ranking
vulnerable to incidental cues such as tokenization length, OCR content, or
text-length priors.

We therefore assign distinct roles to alignment and CPI. Alignment removes
coarse mismatches, while phrase-level controlled-substitution scoring asks how
the surviving captions match their images: whether object, attribute, and
relation phrases measurably affect the image--text score, or whether the match
is sustained by a small number of dominant cues. Concretely, we adopt the top70
subset as the Stage-1 pool $D_{\text{align70}}$ for all subsequent analyses,
because it preserves the favorable coarse properties of alignment filtering
while leaving sufficient sample volume for Stage-2 refinement.
\section{Theoretical Framework: Counterfactual Phrase Intervention}
\label{sec:framework}
 
We formulate CPI as a phrase-level controlled-substitution signal for data
curation. The key idea is to separate coarse pair-level compatibility from the
phrase-level sensitivity that makes a caption useful for compositional
pretraining. We first state this distinction as a curation objective, rather
than as an exact information-theoretic decomposition, then define the
controlled-substitution estimator used for ranking examples.
 
\subsection{Pair-level compatibility and phrase-level sensitivity}
\label{sec:framework:decomposition}

Consider an image-caption pair $(I,T)$ with candidate phrases
$T=(p_1,\ldots,p_K)$. Pair-level curation methods operate on a single global
score $s(I,T)$ and therefore estimate only coarse image-caption compatibility.
Our central premise is that compositional pretraining also depends on how the
measured match responds to the caption's constituent objects, attributes, and
relations.

We posit the following curation objective, rather than an exact
information-theoretic identity:
\begin{equation}
\label{eq:decomposition}
J_{\mathrm{curation}}(I,T)
:=
J_{\mathrm{global}}(I,T)
+
\sum_{j=1}^{K}
J_{\mathrm{phrase}}(I;p_j \mid T_{-j}).
\end{equation}
The first term represents coarse pair-level compatibility, while the conditional
terms represent phrase-level sensitivity within an otherwise fixed caption
context. The objective is intended to organize the data-selection problem: two
captions with similar global compatibility can differ sharply in whether their
objects, attributes, and relations affect the measured image--text score. CPI
does not estimate mutual information or certify grounding; it constructs a
controlled-substitution proxy for ranking examples by relative phrase-level
sensitivity.
 
\subsection{Controlled-substitution estimator}
\label{sec:framework:estimator}

Directly estimating the conditional terms in
Equation~\ref{eq:decomposition} is intractable. CPI therefore uses a controlled
substitution proxy. For each phrase $p_j$ with nominal target token $w_j$, we
construct a perturbed caption
\begin{equation}
\tilde{T}_j = T[w_j \leftarrow r_j],
\label{eq:intervention}
\end{equation}
where $r_j$ is a controlled nonce replacement specified below. We use
``intervention'' operationally to mean a controlled text substitution under a
fixed scoring model, not a causal do-intervention. The phrase attribution score
(PAS) is the resulting drop in image-text similarity:
\begin{equation}
\Delta_j
:=
s(I,T) - s(I,\tilde{T}_j).
\label{eq:pas}
\end{equation}
A larger positive $\Delta_j$ indicates stronger sensitivity of the fixed
image--text scorer to the replaced phrase under this controlled substitution.
Near-zero or negative values indicate that the substitution does not reduce, and
may even increase, the measured similarity under the fixed CLIP scorer. We
therefore interpret PAS as a relative curation signal rather than as an absolute
grounding certificate. Aggregating these drops yields a sample-level signal for
relatively phrase-sensitive supervision.
 
\subsection{The Three-Invariance Replacement Protocol}
\label{sec:framework:invariance}

The validity of $\Delta_j$ depends on whether the replacement $r_j$ removes the
lexical content of $w_j$ without introducing unrelated changes to the caption.
CPI therefore controls three confounders that can otherwise affect CLIP
similarity independently of phrase sensitivity: tokenization length, lexical
semantics, and surface syntactic form.

Concretely, CPI replaces $w_j$ with a deterministically generated nonce token
$r_j$. Let $\#\mathrm{subtok}_{\mathrm{CLIP}}(\cdot)$ denote the number of
OpenAI CLIP-BPE subtokens. The replacement is required to satisfy exact
CLIP-BPE subtoken-count invariance,
\begin{equation}
\#\mathrm{subtok}_{\mathrm{CLIP}}(r_j)
\;=\;
\#\mathrm{subtok}_{\mathrm{CLIP}}(w_j),
\label{eq:tokeninv}
\end{equation}
to have no intended lexical meaning, and to preserve the surface form of $w_j$,
including capitalization, plural marking, and possessive morphology. These
constraints are designed to make the similarity drop $\Delta_j$ reflect the
controlled removal of lexical phrase content rather than artifacts of
tokenization, semantic substitution, or surface-form changes.

We refer to the simultaneous satisfaction of these constraints as the
\textbf{Three-Invariance Replacement Protocol}.
 
\subsection{First-order sensitivity}
\label{sec:framework:sensitivity}

The Three-Invariance protocol is designed to make $\Delta_j$ primarily reflect
the controlled replacement of phrase content rather than incidental replacement
artifacts.

\begin{proposition}[First-order sensitivity under Three-Invariance]
\label{prop:controlled_sensitivity}
Under local smoothness of the CLIP text encoder, if $\tilde{T}_j$ is obtained by
replacing the nominal target $w_j$ of phrase $p_j$ with a substitute $r_j$
satisfying tokenization, lexico-semantic, and syntactic invariance, then
$\Delta_j = s(I,T)-s(I,\tilde{T}_j)$ estimates the first-order sensitivity of
the fixed image--text similarity score to the controlled replacement of $w_j$,
up to residual text-encoder artifacts and higher-order phrase-interaction terms.
\end{proposition}
 
\subsection{Sample-level aggregation}
\label{sec:framework:aggregation}

For curation, we aggregate phrase scores
$D_i=\{\Delta_{ij}\}_{j=1}^{K_i}$ into a normalized mean PAS:
\begin{equation}
\dmean_i = \frac{1}{K_i}\sum_{j=1}^{K_i}\Delta_{ij},
\qquad
\mathrm{Score}_i =
\frac{\dmean_i-\overline{\dmean}}{\sigma_{\dmean}},
\label{eq:agg}
\end{equation}
where $\overline{\dmean}$ and $\sigma_{\dmean}$ are computed over the
alignment-filtered pool. Mean PAS is used as the default CPI score; ablations
against alternative aggregations are reported in Section~\ref{sec:abl:design}.
Because the score is used for ranking within the valid $\Dalign$
pool, its absolute sign should not be read as a binary label of whether a
caption is grounded. Instead, higher mean PAS denotes stronger relative
phrase-level sensitivity under the fixed scoring model. Together, Stage~1
removes coarse mismatches using pair-level alignment, while Stage~2 ranks
globally matched pairs by controlled-substitution sensitivity;
Section~\ref{sec:abl:nonredundancy} verifies that this signal does not reduce to
alignment re-ranking.

\section{Method: Hierarchical Coarse-to-Fine Counterfactual Curation}
\label{sec:method}
 
We now translate CPI into an offline data-curation pipeline. The resulting
subset is passed to standard CLIP, NegCLIP, or CE-CLIP training without changing
the objective, optimizer, or architecture.
 
\subsection{Pipeline overview and Stage-1 alignment filtering}
\label{sec:method:pipeline}

Let $D=\{(I_i,T_i)\}_{i=1}^{N}$ denote the raw corpus. CPI produces $\Dours$ via
two stages:
\begin{equation}
D \xrightarrow{\text{pair-level alignment}} \Dalign
\xrightarrow{\text{phrase-level CPI}} \Dours .
\label{eq:pipeline}
\end{equation}

\begin{figure}[!t]
    \centering
    \includegraphics[width=\linewidth]{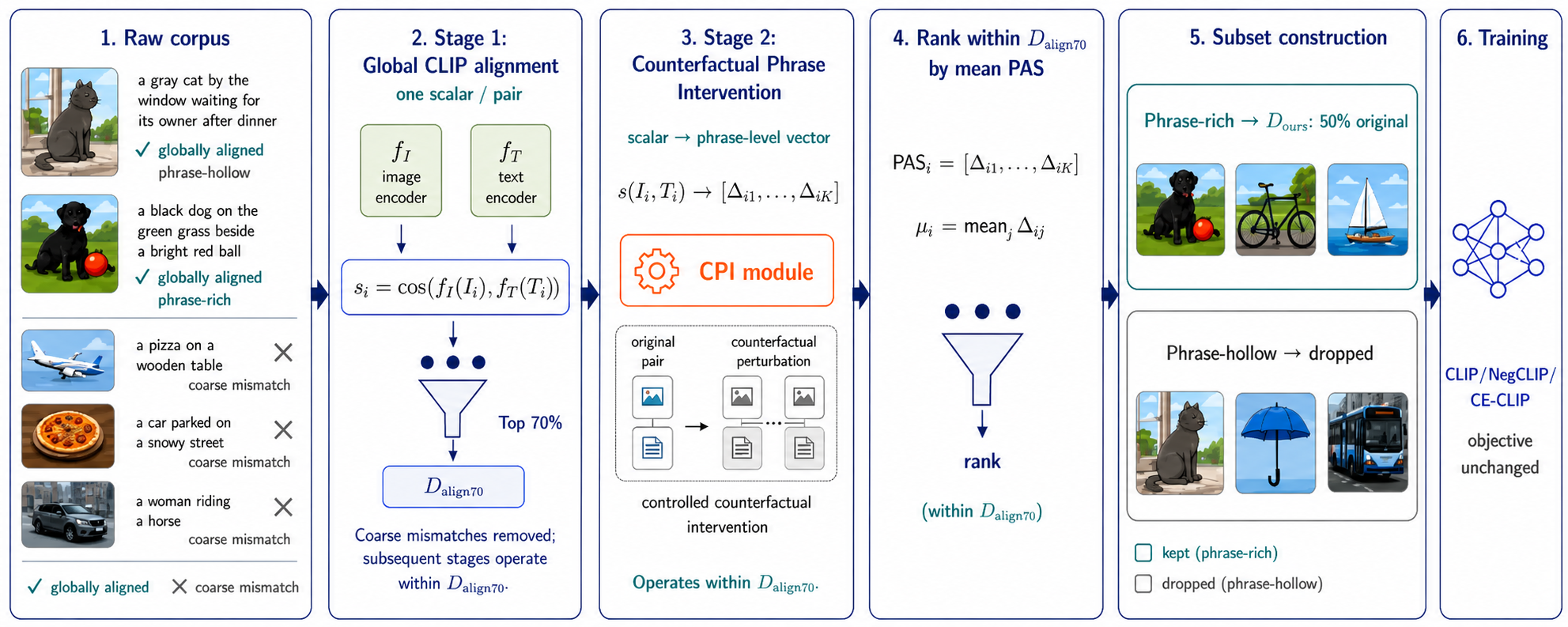}
    \caption{
    Overview of the coarse-to-fine CPI curation pipeline. Stage~1 filters coarse
    mismatches by global CLIP alignment to form $\Dalign$. Stage~2 applies
    controlled phrase substitutions within $\Dalign$, aggregates phrase-level
    similarity drops into mean PAS, and retains the highest-scoring samples as
    $\Dours$ at 50\% of the original corpus. The curated subset is used for
    CLIP, NegCLIP, and CE-CLIP training without modifying the objective.
    }
    \label{fig:overall_pipeline}
\end{figure}

Stage~1 retains the top 70\% of pairs by CLIP similarity, a retention point
chosen from the saturation sweep in Section~\ref{sec:motivation}. Stage~2 then
ranks the surviving high-alignment pool by mean PAS and retains the top-scoring
samples needed to reach 50\% of the original corpus. The two stages play
distinct roles: alignment removes coarse mismatches, while CPI re-ranks globally
matched pairs by relative phrase-level sensitivity under controlled
substitution.

\subsection{Stage~2: Counterfactual Phrase Intervention}
\label{sec:method:stage2}

For each caption in $\Dalign$, we extract candidate object-, relation-, and
predicate-centered phrase spans, select a mutually non-overlapping set, and
identify one nominal target token $w_{ij}$ per span. We then generate a
controlled nonce replacement $r_{ij}$ following the Three-Invariance Replacement
Protocol. Given the perturbed caption
$\tilde{T}_{ij}=T_i[w_{ij}\leftarrow r_{ij}]$, the phrase attribution score
(PAS) is
\begin{equation}
\Delta_{ij}
=
s\bigl(I_i,T_i\bigr)-s\bigl(I_i,\tilde{T}_{ij}\bigr).
\label{eq:delta}
\end{equation}
All PAS values are computed with the same OpenAI CLIP ViT-B/16 encoder used for
Stage~1 alignment, avoiding cross-encoder calibration artifacts.

\begin{figure}[t]
    \centering
    \includegraphics[width=0.95\linewidth]{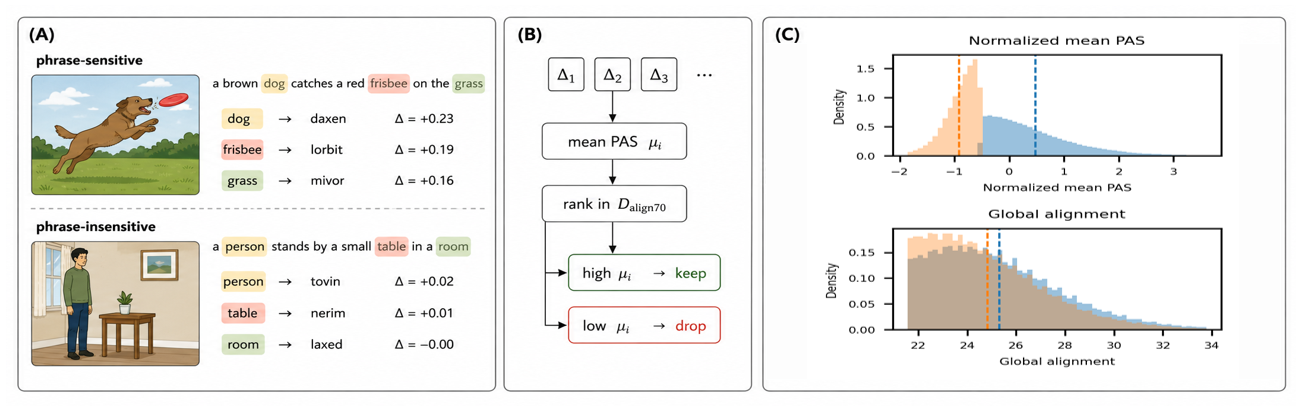}
    \caption{
    Stage~2 CPI scoring: illustration and empirical score distributions.
    \textbf{(A)} Controlled head-to-nonce substitutions for phrase-sensitive and
    phrase-insensitive globally aligned examples.
    \textbf{(B)} Phrase-level similarity drops are aggregated into mean PAS, which
    is used to rank samples within $\Dalign$.
    \textbf{(C)} Empirical kept-versus-dropped score distributions in the scored
    $\Dalign$ pool used for Stage~2 selection. CPI-kept samples are more strongly
    separated from CPI-dropped samples by normalized mean PAS than by global
    alignment, indicating that Stage~2 uses a phrase-level signal that is associated
    with, but not equivalent to, pair-level alignment.
    }
    \label{fig:stage2_cpi}
\end{figure}

Figure~\ref{fig:stage2_cpi} illustrates how CPI converts phrase-level similarity
drops into a sample-level ranking signal. In addition to the illustrative
examples, Panel C reports empirical kept-versus-dropped score distributions:
the separation is sharp in normalized mean PAS but much weaker under global
alignment, visually reinforcing that CPI is not merely re-ranking the Stage-1
pool by alignment.
 
\subsection{Sample-level aggregation and subset construction}
\label{sec:method:subset}

After computing PAS values, we aggregate phrase-level scores using the mean-PAS
sample score in Eq.~\ref{eq:agg}. Samples without valid perturbations are
excluded, affecting only 0.21\% of the processed $\Dalign$ pool. We rank the
remaining samples by mean PAS and retain the top-scoring samples required to
reach 50\% retention relative to the original corpus. Equivalently, Stage~2
drops the bottom $\sim$28.5\% of the valid $\Dalign$ pool by relative
phrase-level sensitivity. The resulting subset is used unchanged in all
subsequent CLIP, NegCLIP, and CE-CLIP training runs.

Mean aggregation favors captions for which multiple constituent phrases affect
the measured image--text score, rather than captions whose score is dominated by
a single cue; Section~\ref{sec:abl:design} validates this choice against
alternative aggregations.
\section{Main Results}
\label{sec:results}

\paragraph{Experimental setting.}
We evaluate CPI on a CC3M-derived corpus of 2{,}222{,}261 image-caption pairs:
Stage~1 retains 1{,}557{,}650 pairs by global alignment, and Stage~2 produces
Ours50 with 1{,}111{,}131 pairs, or 50.0\% of the original corpus. For CLIP
curation comparisons, all runs use CLIP ViT-B/32 from the same checkpoint and
keep optimization, seeds, and objective fixed; only the selected training
examples differ. We compare Full, Full-Random50, Align-only top50, and Ours50,
apply the same subset unchanged to NegCLIP and CE-CLIP, and evaluate on
zero-shot, retrieval, linear-probe, SugarCrepe, SugarCrepe++, and
VL-CheckList-VG benchmarks.
 
\subsection{Two-stage curation improves phrase-sensitive metrics at matched budget}
\label{sec:results:main}
 
We compare Ours50 against Full, Full-Random50, and Align-only top50, covering
no-filtering, data-size, and aggressive pair-level filtering controls
(Tables~\ref{tab:main_general}--\ref{tab:main_compositional}).
 
\begin{table}[t]
\centering
\caption{
Main result---general transfer benchmarks. Values with $\pm$ report mean
$\pm$ standard deviation over two seeds. Retrieval columns average T2I and I2T
scores.
}
\label{tab:main_general}
\footnotesize
\setlength{\tabcolsep}{4pt}
\begin{tabular}{lcccccc}
\toprule
Method & Keep & IN-1K & F R@1 & F R@5 & CIFAR ZS & LP Avg \\
\midrule
Full & 100\% & $7.27{\pm}0.07$ & $12.16{\pm}0.11$ & $30.20{\pm}0.01$ & $29.97$ & $47.30$ \\
Full-Random50 & 50\% & $7.06{\pm}0.00$ & $11.86{\pm}0.28$ & $29.06{\pm}0.25$ & $\mathbf{30.15}$ & $47.39$ \\
Align-only top50 & 50\% & $7.30{\pm}0.00$ & $11.72{\pm}0.70$ & $30.75{\pm}0.38$ & $27.86$ & $47.36$ \\
Ours50 & 50\% & $\mathbf{7.43{\pm}0.01}$ & $\mathbf{12.30{\pm}0.00}$ & $\mathbf{30.94{\pm}0.32}$ & $29.95$ & $\mathbf{47.97}$ \\
\bottomrule
\end{tabular}
\end{table}
 
\begin{table}[t]
\centering
\caption{
Main result---compositional benchmarks. Values report mean $\pm$ standard
deviation over two seeds. We report compact SugarCrepe and SugarCrepe++
summaries together with VL-CheckList-VG aspect results. SC Replace/Add Agg
is computed over the corresponding SugarCrepe replace and add evaluation
examples, and therefore need not equal the unweighted average of the
subset-level accuracies. SC++ Overall follows the stricter acc-both metric.
}
\label{tab:main_compositional}
\footnotesize
\setlength{\tabcolsep}{4.2pt}
\begin{tabular}{lcccc}
\toprule
Metric & Full & Full-Random50 & Align-only top50 & Ours50 \\
\midrule
SC Overall & $57.78 \pm 0.09$ & $57.56 \pm 0.33$ & $57.99 \pm 0.43$ & $\mathbf{58.49 \pm 0.45}$ \\
SC Add Avg & $59.70 \pm 0.18$ & $59.77 \pm 0.60$ & $59.31 \pm 1.18$ & $\mathbf{60.23 \pm 0.03}$ \\
SC Replace/Add Agg & $58.01 \pm 0.31$ & $58.01 \pm 0.37$ & $58.69 \pm 0.05$ & $\mathbf{58.84 \pm 0.11}$ \\
\midrule
SC++ Replace Avg & $47.01 \pm 0.28$ & $45.90 \pm 0.51$ & $47.26 \pm 0.23$ & $\mathbf{47.61 \pm 0.14}$ \\
SC++ Overall (acc-both) & $45.53 \pm 0.30$ & $44.30 \pm 0.01$ & $\mathbf{45.75 \pm 0.31}$ & $45.60 \pm 0.03$ \\
\midrule
VLC Object & $70.68 \pm 0.17$ & $71.15 \pm 0.03$ & $70.78 \pm 0.01$ & $\mathbf{71.29 \pm 0.04}$ \\
VLC Attribute & $56.67 \pm 0.23$ & $56.75 \pm 0.17$ & $57.49 \pm 0.01$ & $\mathbf{57.53 \pm 0.01}$ \\
VLC Relation & $37.71 \pm 0.43$ & $39.15 \pm 0.50$ & $38.62 \pm 0.02$ & $\mathbf{39.62 \pm 0.12}$ \\
VLC Overall & $66.51 \pm 0.11$ & $67.06 \pm 0.04$ & $66.73 \pm 0.01$ & $\mathbf{67.27 \pm 0.04}$ \\
\bottomrule
\end{tabular}
\end{table}

Ours50 yields its strongest gains on phrase-sensitive compositional metrics.
VL-CheckList-VG Relation improves by \textbf{+1.91} over Full and
\textbf{+1.00} over Align-only top50, while SugarCrepe overall improves by
\textbf{+0.71} and \textbf{+0.50}, respectively. Ours50 also improves
SugarCrepe Replace/Add Agg and SugarCrepe++ Replace Avg over the size- and
alignment-based controls; the stricter SugarCrepe++ acc-both overall metric
remains competitive with alignment-only filtering.

The matched-budget controls separate data volume from selection rule:
Full-Random50 tests whether gains arise merely from using half the data, while
Align-only top50 tests whether stricter global alignment alone is sufficient.
Across the nine compact compositional metrics in
Table~\ref{tab:main_compositional}, Ours50 outperforms Full on all nine and
Align-only top50 on 8/9 metrics (sign test against Align-only, $p=0.039$),
indicating that the gain is not driven by a single aggregate score. Importantly,
these compositional gains do not come at the cost of general transfer: Ours50
achieves the strongest ImageNet zero-shot, Flickr R@1/R@5 averages, and
linear-probe average, while remaining competitive on CIFAR zero-shot. The gains are strongest on phrase-sensitive axes reflected by SC Add Avg,
SC Replace/Add Agg, SC++ Replace Avg, and VL-CheckList-VG Relation, consistent
with CPI's controlled-substitution mechanism.
 
\subsection{CPI complements objective-side compositional training}
\label{sec:results:plugin}

By \emph{loss-orthogonal}, we mean that the curated subset is applied without
modifying the training objective, optimizer, or architecture; only the training
data identities differ. CPI is therefore complementary to objective-side methods
such as NegCLIP and CE-CLIP, which modify the learning signal while using the
provided positive pairs. Table~\ref{tab:plugin_compact} evaluates whether the
same CPI-curated 50\%-data subset improves these objectives unchanged.

\begin{table}[t]
\centering
\caption{
Plug-in compatibility with CPI-curated data. Values report mean $\pm$ standard
deviation over two seeds; bold compares Full vs. Ours50 within each objective.
}
\label{tab:plugin_compact}
\footnotesize
\setlength{\tabcolsep}{5pt}
\begin{tabular}{llccccc}
\toprule
Objective & Setting & IN-1K & SC Ovr & SC++ Ovr & VLC Rel & VLC Ovr \\
\midrule
NegCLIP & Full
& $7.05 \pm 0.06$
& $58.91 \pm 0.10$
& $43.46 \pm 0.07$
& $37.72 \pm 1.13$
& $67.46 \pm 0.23$ \\

NegCLIP & \textbf{Ours50}
& $\mathbf{7.15 \pm 0.06}$
& $\mathbf{59.53 \pm 0.62}$
& $\mathbf{43.76 \pm 0.22}$
& $\mathbf{41.56 \pm 0.23}$
& $\mathbf{68.49 \pm 0.14}$ \\
\midrule
CE-CLIP & Full
& $5.41 \pm 0.10$
& $60.20 \pm 0.48$
& $35.46 \pm 0.95$
& $32.72 \pm 0.09$
& $70.24 \pm 0.41$ \\

CE-CLIP & \textbf{Ours50}
& $\mathbf{5.56 \pm 0.06}$
& $\mathbf{61.07 \pm 0.43}$
& $\mathbf{36.60 \pm 0.04}$
& $\mathbf{37.64 \pm 3.38}$
& $\mathbf{70.72 \pm 0.47}$ \\
\bottomrule
\end{tabular}
\end{table}

Replacing only the training data with Ours50 improves the corresponding
full-data counterpart for both methods. The clearest and most stable gain is
NegCLIP on VL-CheckList-VG Relation, from 37.72 to 41.56
(\textbf{+3.84}). CE-CLIP also improves on the same metric, from 32.72 to 37.64
(\textbf{+4.92}), although with larger seed variability ($\pm 3.38$). Thus,
NegCLIP serves as the stable plug-in example, while CE-CLIP provides additional
evidence that data-side phrase sensitivity can complement objective-side
training.

\section{Ablation Studies and Analysis}
\label{sec:abl}

\paragraph{Design ablations.}
\label{sec:abl:design}
In our design diagnostics, mean PAS is the strongest tested aggregation,
reaching 58.77 SugarCrepe overall inside $\Dalign$, compared with 58.70 for the
strongest combined variant and 58.53 for max-only aggregation. A Stage-2
pruning sweep further supports the matched 50\% operating point: moderate
pruning improves compositional accuracy, whereas aggressive pruning degrades
general transfer and compositional performance.

\paragraph{Non-redundancy.}
\label{sec:abl:nonredundancy}
CPI is alignment-aware but not alignment-equivalent: it starts from the globally
aligned pool and reinterprets the image--text match through controlled phrase
substitutions. Within the scored $\Dalign$ pool, normalized mean PAS has only a
weak association with global alignment (Pearson $r=0.084$; Spearman
$\rho=0.088$), indicating that CPI is not simply re-ranking samples by
alignment. Although CPI-kept samples have slightly higher global alignment on
average, the selection rule is controlled-substitution sensitivity rather than
the original alignment score.

We further compare CPI against caption-length and phrase-count heuristics inside
$\Dalign$. These structural proxies are competitive, especially on SugarCrepe++
Overall, but Ours50 remains strongest on ImageNet zero-shot, Flickr R@1 average,
SugarCrepe overall, and VL-CheckList-VG Relation in single-seed diagnostics.
This suggests that CPI captures signal beyond caption verbosity or phrase
density.

Finally, PAS is not reducible to text-side sensitivity or generic nonce OOD
effects: image-conditioned PAS is weakly negatively correlated with text-only
displacement on 50K interventions (Pearson $r=-0.215$, Spearman
$\rho=-0.204$), and nonce drops on 100K interventions lie between
same-category and cross-category real-word controls (mean $\Delta$: 0.0177,
0.0227, 0.0257). These controls support the controlled-substitution
interpretation without making a grounding claim.
\section{Conclusion and Limitations}
\label{sec:conclusion}

We introduced Counterfactual Phrase Intervention (CPI), a coarse-to-fine
data-curation framework that selects globally aligned image-caption pairs by
phrase-level controlled-substitution sensitivity. CPI constructs a 50\%-data
subset without changing the model or objective, improves phrase- and
relation-sensitive metrics while preserving general transfer, and complements
NegCLIP and CE-CLIP as a loss-orthogonal data-side intervention.

Several limitations remain. First, PAS is a relative first-order sensitivity
score, not a grounding, localization, or identification signal: a high score
indicates that controlled phrase replacement changes the measured image--text
similarity, not that the phrase has been visually localized. Second, PAS is
defined relative to the CLIP scoring function used for filtering; whether the
same ranking generalizes to other VLM scoring functions remains open. Third,
CPI relies on rule-based phrase extraction, and nonce replacements may retain
residual text-encoder artifacts. However, our text-only and real-word
calibration controls suggest that PAS is not reducible to these artifacts.
Fourth, our experiments are confined to CC3M-scale CLIP ViT-B/32 with limited
training steps; whether the same trends hold for from-scratch large-scale
pretraining, larger corpora such as CC12M or DataComp, and larger architectures
such as ViT-L/14 is left to future work. Finally, effect sizes are moderate in
absolute scale and several comparisons use two random seeds; we therefore report
standard deviations where available and recommend larger-seed replication. We
also do not provide direct head-to-head comparisons with caption-based or
learned filtering methods such as T-MARS~\citep{maini2023tmars},
DFN~\citep{fang2023data}, and Sieve~\citep{mahmoud2024sieve}; faithful
comparison would require matching corpus scale, model choices, compute budgets,
and operating points across methods, which we leave to future work.

Overall, CPI suggests that compositional VLMs benefit not only from stronger
objectives, but also from selecting positive pairs with stronger measured
phrase-level sensitivity. Its cached-embedding design makes the offline
curation cost amortizable across objectives.


{
\small
\bibliographystyle{unsrtnat}   
\bibliography{reference}}

\begin{thebibliography}{24}
\providecommand{\natexlab}[1]{#1}
\providecommand{\url}[1]{\texttt{#1}}
\expandafter\ifx\csname urlstyle\endcsname\relax
  \providecommand{\doi}[1]{doi: #1}\else
  \providecommand{\doi}{doi: \begingroup \urlstyle{rm}\Url}\fi

\bibitem[Hsieh et~al.(2023)Hsieh, Zhang, Ma, Kembhavi, and Krishna]{hsieh2023sugarcrepe}
Cheng-Yu Hsieh, Jieyu Zhang, Zixian Ma, Aniruddha Kembhavi, and Ranjay Krishna.
\newblock {SugarCrepe}: Fixing hackable benchmarks for vision-language compositionality.
\newblock In \emph{Advances in Neural Information Processing Systems (NeurIPS)}, 2023.

\bibitem[Dumpala et~al.(2024)Dumpala, Jaiswal, Sastry, Milios, Oore, and Sajjad]{dumpala2024sugarcrepe}
Sri~Harsha Dumpala, Aman Jaiswal, Chandramouli Sastry, Evangelos Milios, Sageev Oore, and Hassan Sajjad.
\newblock {SugarCrepe++}: Beyond accuracy in compositional understanding.
\newblock \emph{Advances in Neural Information Processing Systems (NeurIPS)}, 2024.

\bibitem[Hessel et~al.(2021)Hessel, Holtzman, Forbes, Bras, and Choi]{hessel2021clipscore}
Jack Hessel, Ari Holtzman, Maxwell Forbes, Ronan~Le Bras, and Yejin Choi.
\newblock {CLIPScore}: A reference-free evaluation metric for image captioning.
\newblock In \emph{Conference on Empirical Methods in Natural Language Processing (EMNLP)}, pages 7514--7528, 2021.

\bibitem[Gadre et~al.(2023)Gadre, Ilharco, Fang, Hayase, Smyrnis, Nguyen, Marten, Wortsman, Ghosh, Zhang, et~al.]{gadre2023datacomp}
Samir~Yitzhak Gadre, Gabriel Ilharco, Alex Fang, Jonathan Hayase, Georgios Smyrnis, Thao Nguyen, Ryan Marten, Mitchell Wortsman, Dhruba Ghosh, Jieyu Zhang, et~al.
\newblock {DataComp}: In search of the next generation of multimodal datasets.
\newblock In \emph{Advances in Neural Information Processing Systems (NeurIPS)}, 2023.

\bibitem[Maini et~al.(2024)Maini, Goyal, Lipton, Kolter, and Raghunathan]{maini2023tmars}
Pratyush Maini, Sachin Goyal, Zachary~C. Lipton, J.~Zico Kolter, and Aditi Raghunathan.
\newblock {T-MARS}: Improving visual representations by circumventing text feature learning.
\newblock In \emph{International Conference on Learning Representations (ICLR)}, 2024.
\newblock Originally arXiv:2307.03132, 2023.

\bibitem[Fang et~al.(2024)Fang, Jose, Jain, Schmidt, Toshev, and Shankar]{fang2023data}
Alex Fang, Albin~Madappally Jose, Amit Jain, Ludwig Schmidt, Alexander Toshev, and Vaishaal Shankar.
\newblock Data filtering networks.
\newblock In \emph{International Conference on Learning Representations (ICLR)}, 2024.
\newblock arXiv:2309.17425, 2023.

\bibitem[Mahmoud et~al.(2024)Mahmoud, Elhoushi, Abbas, Yang, Ardalani, Leather, and Morcos]{mahmoud2024sieve}
Anas Mahmoud, Mostafa Elhoushi, Amro Abbas, Yu~Yang, Newsha Ardalani, Hugh Leather, and Ari Morcos.
\newblock {Sieve}: Multimodal dataset pruning using image captioning models.
\newblock In \emph{Proceedings of the IEEE/CVF Conference on Computer Vision and Pattern Recognition (CVPR)}, 2024.

\bibitem[Xu et~al.(2024)Xu, Xie, Tan, Huang, Howes, Sharma, Li, Ghosh, Zettlemoyer, and Feichtenhofer]{xu2024demystifying}
Hu~Xu, Saining Xie, Xiaoqing~Ellen Tan, Po-Yao Huang, Russell Howes, Vasu Sharma, Shang-Wen Li, Gargi Ghosh, Luke Zettlemoyer, and Christoph Feichtenhofer.
\newblock Demystifying {CLIP} data.
\newblock In \emph{International Conference on Learning Representations (ICLR)}, 2024.

\bibitem[Udandarao et~al.(2025)Udandarao, Parthasarathy, Naeem, Evans, Albanie, Tombari, Xian, Tonioni, and H{\'e}naff]{udandarao2025active}
Vishaal Udandarao, Nikhil Parthasarathy, Muhammad~Ferjad Naeem, Talfan Evans, Samuel Albanie, Federico Tombari, Yongqin Xian, Alessio Tonioni, and Olivier~J. H{\'e}naff.
\newblock Active data curation effectively distills large-scale multimodal models.
\newblock In \emph{Proceedings of the IEEE/CVF Conference on Computer Vision and Pattern Recognition (CVPR)}, 2025.

\bibitem[Tirumala et~al.(2023)Tirumala, Simig, Aghajanyan, and Morcos]{tirumala2023d4}
Kushal Tirumala, Daniel Simig, Armen Aghajanyan, and Ari Morcos.
\newblock {D4}: Improving {LLM} pretraining via document de-duplication and diversification.
\newblock In \emph{Advances in Neural Information Processing Systems (NeurIPS)}, 2023.

\bibitem[Abbas et~al.(2023)Abbas, Tirumala, Simig, Ganguli, and Morcos]{abbas2023semdedup}
Amro Abbas, Kushal Tirumala, D{\'a}niel Simig, Surya Ganguli, and Ari~S. Morcos.
\newblock {SemDeDup}: Data-efficient learning at web-scale through semantic deduplication.
\newblock In \emph{ICLR Workshop on Multimodal Representation Learning}, 2023.

\bibitem[Mindermann et~al.(2022)Mindermann, Brauner, Razzak, Sharma, Kirsch, Xu, H{\"o}ltgen, Gomez, Morisot, Farquhar, and Gal]{mindermann2022prioritized}
S{\"o}ren Mindermann, Jan~M. Brauner, Muhammed~T. Razzak, Mrinank Sharma, Andreas Kirsch, Winnie Xu, Benedikt H{\"o}ltgen, Aidan~N. Gomez, Adrien Morisot, Sebastian Farquhar, and Yarin Gal.
\newblock Prioritized training on points that are learnable, worth learning, and not yet learnt.
\newblock In \emph{Proceedings of the 39th International Conference on Machine Learning (ICML)}, pages 15630--15649, 2022.

\bibitem[Evans et~al.(2024)Evans, Pathak, Merzic, Schwarz, Khetarpal, Tang, Markeeva, Hill, and Pascanu]{evans2024bad}
Talfan Evans, Shreya Pathak, Hamza Merzic, Jonathan Schwarz, Khimya Khetarpal, Jordan Tang, Larisa Markeeva, Felix Hill, and Razvan Pascanu.
\newblock Bad students make great teachers: Active learning accelerates large-scale visual understanding.
\newblock In \emph{Advances in Neural Information Processing Systems (NeurIPS)}, 2024.

\bibitem[Yuksekgonul et~al.(2023)Yuksekgonul, Bianchi, Kalluri, Jurafsky, and Zou]{yuksekgonul2023when}
Mert Yuksekgonul, Federico Bianchi, Pratyusha Kalluri, Dan Jurafsky, and James Zou.
\newblock When and why vision-language models behave like bags-of-words, and what to do about it?
\newblock In \emph{International Conference on Learning Representations (ICLR)}, 2023.

\bibitem[Zhang et~al.(2024)Zhang, Awal, and Agrawal]{zhang2024ce}
Le~Zhang, Rabiul Awal, and Aishwarya Agrawal.
\newblock Contrasting intra-modal and ranking cross-modal hard negatives to enhance visio-linguistic compositional understanding.
\newblock In \emph{Proceedings of the IEEE/CVF Conference on Computer Vision and Pattern Recognition (CVPR)}, 2024.

\bibitem[Oh et~al.(2024)Oh, Lim, Kim, Han, Yun, Choo, Hauptmann, Cheng, and Song]{oh2024fsc}
Changdae Oh, Hyesu Lim, Mijoo Kim, Dongyoon Han, Sangdoo Yun, Jaegul Choo, Alexander Hauptmann, Zhi-Qi Cheng, and Kyungwoo Song.
\newblock Preserving multi-modal capabilities of pre-trained {VLM}s for improving vision-linguistic compositionality.
\newblock In \emph{Conference on Empirical Methods in Natural Language Processing (EMNLP)}, 2024.

\bibitem[Patel et~al.(2024)Patel, Kusumba, Cheng, Kim, Gokhale, Baral, and Yang]{patel2024triplet}
Maitreya Patel, Abhiram Kusumba, Sheng Cheng, Changhoon Kim, Tejas Gokhale, Chitta Baral, and Yezhou Yang.
\newblock {TripletCLIP}: Improving compositional reasoning of {CLIP} via synthetic vision-language negatives.
\newblock In \emph{Advances in Neural Information Processing Systems (NeurIPS)}, 2024.

\bibitem[Doveh et~al.(2023)Doveh, Arbelle, Harary, Herzig, Kim, Cascante-Bonilla, Alfassy, Panda, Giryes, Feris, et~al.]{doveh2024dense}
Sivan Doveh, Assaf Arbelle, Sivan Harary, Roei Herzig, Donghyun Kim, Paola Cascante-Bonilla, Amit Alfassy, Rameswar Panda, Raja Giryes, Rogerio Feris, et~al.
\newblock Dense and aligned captions ({DAC}) promote compositional reasoning in {VL} models.
\newblock In \emph{Advances in Neural Information Processing Systems (NeurIPS)}, 2023.

\bibitem[Sahin et~al.(2024)Sahin, Li, Khan, Cremers, and Tresp]{sahin2024read}
Ugur Sahin, Hang Li, Qadeer Khan, Daniel Cremers, and Volker Tresp.
\newblock {READ}: Reinforcement-based adversarial learning for text-image compositional reasoning.
\newblock In \emph{Proceedings of the IEEE/CVF Winter Conference on Applications of Computer Vision (WACV)}, 2024.

\bibitem[Kim et~al.(2018)Kim, Wattenberg, Gilmer, Cai, Wexler, Viegas, and Sayres]{kim2018interpretability}
Been Kim, Martin Wattenberg, Justin Gilmer, Carrie Cai, James Wexler, Fernanda Viegas, and Rory Sayres.
\newblock Interpretability beyond feature attribution: Quantitative testing with concept activation vectors ({TCAV}).
\newblock In \emph{Proceedings of the 35th International Conference on Machine Learning (ICML)}, pages 2668--2677, 2018.

\bibitem[Sundararajan et~al.(2017)Sundararajan, Taly, and Yan]{sundararajan2017axiomatic}
Mukund Sundararajan, Ankur Taly, and Qiqi Yan.
\newblock Axiomatic attribution for deep networks.
\newblock In \emph{Proceedings of the 34th International Conference on Machine Learning (ICML)}, pages 3319--3328, 2017.

\bibitem[Oikarinen and Weng(2023)]{oikarinen2023clip}
Tuomas Oikarinen and Tsui-Wei Weng.
\newblock {CLIP-Dissect}: Automatic description of neuron representations in deep vision networks.
\newblock In \emph{International Conference on Learning Representations (ICLR)}, 2023.

\bibitem[Meng et~al.(2022)Meng, Bau, Andonian, and Belinkov]{meng2022locating}
Kevin Meng, David Bau, Alex~J. Andonian, and Yonatan Belinkov.
\newblock Locating and editing factual associations in {GPT}.
\newblock In \emph{Advances in Neural Information Processing Systems (NeurIPS)}, 2022.

\bibitem[Ettinger(2020)]{ettinger2020what}
Allyson Ettinger.
\newblock What {BERT} is not: Lessons from a new suite of psycholinguistic diagnostics for language models.
\newblock \emph{Transactions of the Association for Computational Linguistics (TACL)}, 8:\penalty0 34--48, 2020.

\end{thebibliography}




\end{document}